\documentclass{ifacconf}

\usepackage{graphicx}      % include this line if your document contains figures
\usepackage{natbib}        % required for bibliography
\usepackage{amsmath,amsfonts}
\usepackage{dsfont}
\usepackage{xcolor}
\usepackage{bm}
\usepackage{url}
\usepackage{siunitx}
\usepackage{comment}

\makeatletter
\let\old@ssect\@ssect % Store how ifacconf defines \@ssect
\makeatother

\usepackage{natbib}
\usepackage{hyperref}

\makeatletter
\def\@ssect#1#2#3#4#5#6{%
  \NR@gettitle{#6}% Insert key \nameref title grab
  \old@ssect{#1}{#2}{#3}{#4}{#5}{#6}% Restore ifacconf's \@ssect
}
\makeatother

\newcommand{\zak}[1]{({\color{magenta}Zak: #1})}

\newtheorem{remark}[thm]{Remark}
\newtheorem{proposition}{Proposition}

\begin{document}
\begin{frontmatter}

\title{Physically Consistent Neural ODEs for Learning Multi-Physics Systems\thanksref{footnoteinfo}} 
%\title{Irreversible port-Hamiltonian Neural ODEs to Learn Multi-physics Systems\thanksref{footnoteinfo}} 

\thanks[footnoteinfo]{This research was supported by the Swiss National Science Foundation under the NCCR Automation (grant agreement 51NF40\_180545) and in part by the Swiss Data Science Center (grant no. C20-13). $^\dagger$The authors contributed equally. Corresponding authors: \url{muhammad.zakwan@epfl.ch}, \url{loris.dinatale@empa.ch}.}

\author[1]{M. Zakwan$^\dagger$}
\author[2,1]{L. Di Natale$^\dagger$} 
\author[2]{B. Svetozarevic} 
\author[2]{P. Heer}
\author[1]{C. N. Jones}
\author[1]{G. Ferrari Trecate} 

\address[1]{Laboratoire d'Automatique, EPFL, Lausanne, Switzerland.}
\address[2]{Urban Energy Systems Laboratory, Empa, D\"{u}bendorf, Switzerland.}
%\address[2]{Urban Energy Systems Laboratory, Swiss Federal Laboratories for Material Science (Empa), 8600 Duebendrof, Switzerland.} 

\begin{abstract}                % Abstract of not more than 250 words.
    Despite the immense success of neural networks %and classical data-driven methods 
    in modeling system dynamics from data, they often remain physics-agnostic black boxes. In the particular case of physical systems, they might consequently make physically inconsistent predictions, which makes them unreliable in practice. In this paper, we leverage the framework of Irreversible port-Hamiltonian Systems (IPHS), which can describe most multi-physics systems, and rely on Neural Ordinary Differential Equations (NODEs) to learn their parameters from data. Since IPHS models are consistent with the first and second principles of thermodynamics by design, so are the proposed Physically Consistent NODEs (PC-NODEs). %, i.e. the conservation of energy and the irreversible creation of entropy.
    Furthermore, the NODE training procedure allows us to seamlessly incorporate prior knowledge of the system properties in the learned dynamics.
    %and can seamlessly incorporate prior knowledge to ease the learning procedure. 
    We demonstrate the effectiveness of the proposed method by learning the thermodynamics of a building from the real-world measurements and the dynamics of a simulated gas-piston system. Thanks to the modularity and flexibility of the IPHS framework, PC-NODEs can be extended to learn physically consistent models of multi-physics distributed systems.

%they can be physically inconsistent, notably in modeling thermodynamic systems. Therefore, it may pose difficulties while designing controllers and observers. In this paper, based on irreversible port-Hamiltonian dynamics, we propose Neural Ordinary Differential Equations (NODEs) – a family of continuous-depth NNs represented by dynamical systems - to model thermodynamic systems. By construction, the proposed NODEs are consistent with the first and second principles of thermodynamics, i.e., conservation of energy and irreversible creation of entropy. Moreover, our framework allows one to incorporate priors, multi-physics systems (e.g., electromechanical), model distributed systems, and design controllers by interconnection with other passive port-Hamiltonian systems. We demonstrate the efficacy of our NODEs by thermodynamic modeling of a real-world building and a simulated gas-piston system.    
\end{abstract}

\end{frontmatter}
%===============================================================================

\section{Introduction}
% Paragraph no. 1
% Motivation of using neural networks for systems identification and cite some references 
% Issues with common neural networks, that is, lack of physical consistency, these model cannot be used for controller design.
% Other physics-inspired networks such as lagrangian nets, hamiltonian nets, PCNNs. 

%  i) Our key message is to provide a neural ODE that is physically consistent as opposed to Black box methods, e.g. LSTMs, and ARX. 
%                      ii) Key benefits of irreversible port Hamiltonian systems, such as modeling multi-physics, structural properties, controller design, and interconnection with other port-Hamiltonian systems. 
%                      iii) A brief literature survey to strengthen the idea of physics-based NNs. 
                     
In recent years, Neural Networks (NNs) have achieved impressive performances on a broad range of tasks, including time series prediction, where Recurrent NNs (RNNs), Gated Recurrent Units (GRUs), Long Short-Term Memory networks (LSTMs), and transformers, often attain great accuracy \citep{wang2021physics}. 
These successes also motivated researchers to use NNs to identify system dynamics from data, but such models %, replacing classical methods such as AutoRegressive models with eXogenours inputs (ARX). 
%However, purely data-driven methods, typically the ones based on NNs, 
often suffer from physical inconsistencies: they can fit data well without learning the underlying ground truth, making them unreliable in practice~\citep{geirhos2020shortcut, di2022physically}. 
As a countermeasure to the brittleness of NN-based models, there has been increasing interest in incorporating prior knowledge -- also known as inductive bias -- into NNs to ensure physical consistency, leading to Hamiltonian NNs \citep{chen2019symplectic, greydanus2019hamiltonian, finzi2020simplifying}, Lagrangian NNs \citep{cranmer2020lagrangian}, or Poisson NNs \citep{jin2022learning}, amongst others.
We defer the reader to \citet{wang2021physics} for a comprehensive survey on physics-guided deep learning for dynamical systems.

%InNeural Networks (NNs) have demonstrated remarkable performance in learning patterns from data in many different contexts, for instance, learning the underlying dynamics from time-series data. Several state-of-the-art black-box models have been proposed to fit sequential data, for example, recurrent neural nets (RNNs), gated recurrent units, long-short term memory (LSTMs) networks, and transformers. 
%Moreover, classical grey-box methods such as ARX and nonlinear ARX are also available for system identification. However, these methods yield models that may not be stable or are not necessarily consistent with the physics of the underlying systems. It becomes the main caveat while using these models to design controllers and observers for the predictions.  
%Recently, there has been an increasing interest in applying  NNs to learn physical models from data by incorporating prior knowledge, for example, Hamiltonian NNs \citep{chen2019symplectic, greydanus2019hamiltonian, finzi2020simplifying}, Lagrangian NNs \citep{cranmer2020lagrangian}, and Poisson NNs \citep{jin2022learning}. We defer the reader to \citet{wang2021physics} for a comprehensive survey on physics-guided deep learning for dynamical systems.

At the same time, higher-level connections between NNs and dynamical systems have also been studied, showing that some classes of NNs can be interpreted as discretized dynamical systems~\citep{RN10726}.
On the other hand, \citet{chen2018neural} proposed the framework of Neural Ordinary Differential Equations (NODEs), where  inputs are transformed through a continuous-time ODE embedding trainable parameters. 
In other words, NODEs learn the parameters of an ODE to fit data, making them particularly suitable to model complex dynamical systems~\citep{greydanus2019hamiltonian, RN11780}. 
Furthermore, their interpretation as ODEs allows one to borrow tools from dynamical system theory to analyze their properties~\citep{zakwan2022Robust, fazlyabSafetyVerificationRobustness2022, galimberti2021hamiltonian}. 
However, similarly to classical NNs, NODEs can be physically inconsistent in general.

%Recently, the connections between NNs and dynamical systems have been extensively explored.
%Representative results include classes of NNs stemming from the discretization of dynamical systems~\citep{RN10726} and  NODEs~\citep{chen2018neural},  which transform the input through a continuous-time ODE embedding training parameters. 
%The continuous-time nature of NODEs makes them particularly suitable for learning complex dynamical systems~\citep{greydanus2019hamiltonian, RN11780} and allows borrowing tools from dynamical system theory to analyze their properties~\citep{zakwan2022Robust, fazlyabSafetyVerificationRobustness2022, RN11564}. However, like NNs, NODEs can also be physically inconsistent. 

This paper proposes %a framework unifying both worlds and creating 
Physically Consistent NODEs (PC-NODEs), which leverage the Irreversible port-Hamiltonian (IPH) modeling framework to describe multi-physics systems and %the effectiveness of 
NODEs to learn their dynamics. Thanks to the IPH formulation, we can guarantee that PC-NODEs respect the %underlying physical laws at all times and by construction, typically the
first and second laws of thermodynamics at all times and by construction, solving the issue of physically inconsistent NODEs. Moreover, unlike black-box NNs, %models, % such as RNNs and LSTMs, 
PC-NODEs allow us to embed \textit{a priori} desired structural properties of trainable parameters, such as skew-symmetry and prescribed sparsity patterns, in the learning process.

Our efforts to ground learning schemes in the underlying physics are conceptually related to the work of \cite{masi2021thermodynamics}, who %encoded the two principles of thermodynamics in NNs by %leveraging automatic differentiation to 
ensured that suitable model derivatives are consistent with the rules of thermodynamics. 
In another attempt, \cite{di2022physically} introduced Physically Consistent NNs (PCNNs), where a physics-inspired module runs in parallel to a NN to ensure the predictions comply with underlying physical laws. 
While these methods were shown to work well in case studies, they are limited to a few applications. 
In contrast, PC-NODEs are more general and applicable to a wide variety of systems.

The modularity of the IPH framework allows us to characterize many multi-physics systems, including  % such as Continuously Stirred Tank Reactors (CSTRs), and many 
thermodynamic, mechanical, chemical, or electrical systems~\citep{ramirez2013irreversible, van2021energy}. 
Furthermore, identifying system dynamics in the IPH form provides several benefits, as one can then design stabilizing controllers and scale to distributed systems via interconnection with other passive port-Hamiltonian systems~\citep{ramirez2013irreversible}. To showcase the flexibility of the proposed PC-NODEs, in this paper, we model the thermal dynamics of a building from the real-world measurements and the dynamics of a simulated gas-piston system.

{\it Organization:} Section~2 presents PC-NODEs and describes the training procedure. In Section~3, we consider the  modeling of two case studies, and the results are illustrated in Section~4. Finally, Section~5 concludes the paper.  

{\it Notations:} 
% The set of non-negative real numbers is $\mathbb{R}^+$. 
% The operator $\frac{\partial f({x})}{\partial {x}}$ denotes the Jacobian of a continuously differentiable function $f(\cdot)$. 
% The maximal and minimal eigenvalues of a matrix $\bm{A}$ are denoted as $\bar{\lambda}(\bm{A})$ and $\underline{\lambda}(\bm{A})$, respectively. 
% $\text{diag}(\bm{x})$ represents a diagonal matrix with the entries of the vector $\bm{x}$ on the diagonal. 
% For symmetric matrices ${A}$ and ${B}$, ${A} \succ (\succeq) {B}$ means that ${A}-{B}$ is positive (semi)definite. 
% The 2-norm is denoted as $||\cdot||$. 
The $p-$norm is denoted as $||\cdot||_p$. A matrix $J$ is \textit{skew-symmetric} if $J=-J^\top$. 
The \emph{Poisson bracket} of $Z,G\in\mathcal{C}^\infty(\mathbb{R}^n)$ with respect to a skew-symmetric matrix $J$ is defined as %$\{\cdot,\cdot\}_J:\mathcal{C}^\infty(\mathbb{R}^n) \times \mathcal{C}^\infty(\mathbb{R}^n) \to \mathcal{C}^\infty(\mathbb{R}^n)$, 
$\{Z, G\}_J = \frac{\partial Z^\top(x)}{\partial x} J \frac{\partial G(x)}{\partial x}$.
%The \emph{Poisson bracket} with respect to a skew-symmetric matrix $J$ defines a map from the pairs of $\mathcal{C}^\infty(\mathbb{R}^n)$ functions $Z$
%and $G$ to a $\mathcal{C}^\infty(\mathbb{R}^n)$ function and is  denoted by $\{Z, G\}_J$ and defined as 
%$\{Z, G\}_J = \frac{\partial Z^\top(x)}{\partial x} J \frac{\partial G(x)}{\partial x}$.
  %for $p = 1,2, \hdots, \infty$. 

\section{Learning Irreversible port-Hamiltonian dynamics}

%  i) Problem statement
%                       ii) We have a complicated physics-inspired neural ODE (PC-NODE). It is consistent with the first and second laws of thermodynamics, and monotone with respect to inputs.
%                       iii) A simplified version to model linear systems and show that this is equivalent to the well-known RC model.
%                       iv) A remark that we can put structure into A and B matrices. 
%                       v) Training procedure  

This section introduces the PC-NODE framework to learn system dynamics from data while ensuring compatibility with the first and second laws of thermodynamics. %This is hardwired in the model thanks to the underlying IPH system formulation.
%The underlying dynamics of NODEs are based on irreversible port Hamiltonian systems \citep{ramirez2013modelling} and therefore are consistent with the first and second principles of
%thermodynamics for all the trainable parameters. Moreover, the structure matrices are
%modulated by a non-linear function that precisely expresses the irreversibility of the system.
%  Neural ODEs are deep neural networks that stem from the discretization of dynamical systems. In this framework, each layer of the neural network corresponds to a discretization step of underlying dynamical system. 

\subsection{Physics-Consistent NODEs}

An IPH system~\citep{ramirez2013irreversible, ramirez2013modelling} is described as %NODE (IPH-NODE) is given as 
\begin{align}
    \label{PC-NODE}
    \dot{x} &= \textcolor{blue}{R} \left(x,\frac{\partial \textcolor{blue}{H}(x)}{\partial x}, \frac{\partial \textcolor{blue}{S}(x)}{\partial x} \right) \textcolor{blue}{J}  \frac{\partial \textcolor{blue}{H}(x)}{\partial x} \nonumber \\
    &\qquad+  \textcolor{blue}{W}\left(x,\frac{\partial \textcolor{blue}{H}(x)}{\partial x} \right) + \textcolor{blue}{g} \left(x,\frac{\partial \textcolor{blue}{H}(x)}{\partial x} \right) u \; ,
\end{align}
where $x \in \mathbb{R}^n$ is the state, $u \in \mathbb{R}^m$ the control input, and the different functions and matrices satisfy\footnote{To have concise notation throughout the paper, the dependence on $x$ and partial derivatives is dropped when it is clear from the context.}%To have concise notation, the dependence on $x$ and partial derivatives is dropped. For the sake of brevity, throughout the paper, we drop the dependence on $x$ and partial derivatives when it is clear.}:%from the functions $\textcolor{blue}{H}(x)$, $\textcolor{blue}{S}(x)$.}:
\begin{itemize}
    \item[$(P_1)$] the Hamiltonian function $\textcolor{blue}{H}$ and the entropy function $\textcolor{blue}{S}$ are maps from $\mathcal{C}^\infty(\mathbb{R}^n)$ to $\rightarrow \mathbb{R}$;
    \item[$(P_2)$] the interconnection matrix $\textcolor{blue}{J} \in \mathbb{R}^{n \times n}$ is constant and skew-symmetric; 
    \item[$(P_3)$] the real function $\textcolor{blue}{R} = \textcolor{blue}{R}(x,\frac{\partial \textcolor{blue}{H}}{\partial x}, \frac{\partial \textcolor{blue}{S}}{\partial x})$ is defined as
    \begin{align}
        \textcolor{blue}{R}\left(x,\frac{\partial \textcolor{blue}{H}}{\partial x}, \frac{\partial \textcolor{blue}{S}}{\partial x} \right) = \textcolor{blue}{\gamma}\left(x, \frac{\partial \textcolor{blue}{H}}{\partial x} \right) \{ \textcolor{blue}{S}, \textcolor{blue}{H} \}_{\textcolor{blue}{J}} \; , \label{equ:R}
    \end{align}
    where $\textcolor{blue}{\gamma} \succeq 0$ is a %non-linear
    nonnegative function of the states and co-states of the system; % that may be expressed as a function of the states only; 
    \item[$(P_4)$] the two vector fields $\textcolor{blue}{W}$ and $\textcolor{blue}{g}$ satisfy $\textcolor{blue}{W}(x,\frac{\partial \textcolor{blue}{H}}{\partial x}) \in \mathbb{R}^n$ and $\textcolor{blue}{g}(x,\frac{\partial \textcolor{blue}{H}}{\partial x}) \in \mathbb{R}^{n \times m}$. % and are associated with the ports of the system.
\end{itemize}

% Remarkably, 
We have used the %the real functions indicated in%
\textcolor{blue}{\bf blue color} to denote functions that can be parameterized, e.g. using NNs, and identified from data as described in Section~\ref{sec:training}. %using NODEs. 
%which gives rise to the proposed PC-NODEs. 
This defines the overall PC-NODE framework. As long as the learned parameters respect the constraints and properties listed above, the learned model will obey the first and second laws of thermodynamics by construction. Indeed, by the skew-symmetry of $\textcolor{blue}{J}$, setting $\textcolor{blue}{W}, u \equiv 0$, we have
\begin{align}
\frac{d \textcolor{blue}{H}}{d t}=\frac{\partial \textcolor{blue}{H}^{\top}}{\partial x}\left(\textcolor{blue}{R} \textcolor{blue}{J} \frac{\partial \textcolor{blue}{H}}{\partial x}\right)
=\textcolor{blue}{R} \times \left(\frac{\partial \textcolor{blue}{H}^{\top}}{\partial x} \textcolor{blue}{J} \frac{\partial \textcolor{blue}{H}}{\partial x}\right)
=0 \;, \label{equ:conservation energy}
\end{align}
which proves the conservation of energy in the system. Similarly, we can show the irreversible creation of entropy in the system as follows:
\begin{align*}
\frac{d \textcolor{blue}{S}}{d t}= \textcolor{blue}{R} \frac{\partial \textcolor{blue}{S}^{\top}}{\partial x}\textcolor{blue}{J} \frac{\partial \textcolor{blue}{H}}{\partial x}
=  \textcolor{blue}{\gamma}\left(x, \frac{\partial \textcolor{blue}{H}}{\partial x} \right) \{\textcolor{blue}{S}, \textcolor{blue}{H} \}_{\textcolor{blue}{J}}^2 \geq 0 \;,
\end{align*}
as long as $\textcolor{blue}{\gamma} \succeq 0$. We defer the reader to \cite{ramirez2013modelling} for more details on these computations. % For more details see \citep{ramirez2013modelling}. 

\subsection{Training PC-NODEs}
\label{sec:training}

Several NODE training procedures have been proposed in the literature, such as the adjoint sensitivity method~\citep{chen2018neural} or the auto-differentiation technique~\citep{paszke2017automatic}. In this work, inspired by \cite{RN10726}, we first discretize PC-NODE~\eqref{PC-NODE} using the Forward-Euler (FE) method with sampling period $h>0$, leading to
\begin{align}
\label{eq:FE}
    {x}_{i + 1 } = {x}_i + h \left( \textcolor{blue}{R} \textcolor{blue}{J}  \frac{\partial \textcolor{blue}{H}(x_i)}{\partial x_i} 
    +  \textcolor{blue}{W}+ \textcolor{blue}{g} u_i \right) ,
\end{align}
where ${x}_i$ and ${x}_{i+1}$ represent the current and next state, respectively. In practice, the step-size $h$ is chosen sufficiently small so as to interpret the states in~\eqref{eq:FE} as a sampled version of the state ${x}(t)$ of system~\eqref{PC-NODE}.

\begin{comment}
\label{remark_discretization}
Several methods have been proposed for training NODEs, such as the adjoint sensitivity method \citep{chen2018neural}, and the auto-differentiation technique \citep{paszke2017automatic}.
In this paper, to specify a NN architecture for the training, we can discretize the IPH-NODE \eqref{PC-NODE}
with a sampling period $h = \frac{T}{L}$, $L \in  \mathbb{N}$ \citep{RN10726}. 
% The $L$ resulting discrete-time equations can be utilized to define the layers of a NN of depth $L$ \loris{Nope, I still disagree, it's not a NN of depth L, just one layer applied recursively, both are \textbf{NOT} the same thing.} \citep{chen2018neural}. 
For instance, using Forward Euler (FE) discretization with a step size of $h > 0$, one obtains 
\begin{align}
\label{eq:FE}
    {x}_{i + 1 } = {x}_i + h \left( \textcolor{blue}{R} \textcolor{blue}{J}  \frac{\partial \textcolor{blue}{H}(x_i)}{\partial x_i} 
    +  \textcolor{blue}{W}+ \textcolor{blue}{g} u_i \right) ,
\end{align}
for $i = 0,1,\cdots, L-1$, where ${x}_i$ and ${x}_{i+1}$ represent the current and next state, respectively. In practice, the step-size $h$ is chosen sufficiently small so as to interpret the states in~\eqref{eq:FE} as a sampled version of the state ${x}(t)$ in~\eqref{PC-NODE}.
\end{comment}

%To this end, we presented the framework of IPH-NODEs and discussed the associated properties. Now we will present the training procedure of NODE \eqref{PC-NODE}. Assume we have the following data-set consisting of $M$ trajectories:
We then assume to have access to a dataset of $M$ sampled full-state trajectories
\begin{align*}
    \mathcal{D}:= \bigg\{(z_0^j ,r_0^j), (z_1^j,r_1^j), \cdots (z_{L}^j,r_{L}^j) \bigg\}_{j = 1}^{M}\; ,
\end{align*}
where $L$ is the total number of time steps for each trajectory of measured states $z$ and inputs $r$. %Moreover, $z$ is the vector comprising of measured thermodynamic quantities, such as temperature, pressure, or volume, and $r$ is the vector of all inputs, i.e., heating elements, external forces, or ambient temperature. Then we have the following optimization problem:
%This allows us to setup the following optimization problem to train PC-NODE~\eqref{eq:FE}
Finally, we train system~\eqref{eq:FE} to minimize the following objective
\begin{align}
\label{obj}
    \min_{\textcolor{blue}{R},\textcolor{blue}{J},\textcolor{blue}{W},\textcolor{blue}{g}} \sum_{i = 1}^L \sum_{j = 1}^M \ell(z_i^j, x_i^j)  \;. %||z_i^j - x_i^j||^{2}_2  \;.  %+ \lambda ||\textcolor{blue}{J}||_1 \;, \\
    %\text{s.t.}\quad \textcolor{blue}{J} = -\textcolor{blue}{J}^\top, \textcolor{blue}{R} \succeq 0\;, \nonumber 
\end{align}
%where $x$ is the output of the NODE \eqref{PC-NODE} and $\lambda >0$ is the regularization weight. 
%where the constraints ensure $\textcolor{blue}{R}$ and $\textcolor{blue}{J}$ have the desired characteristics to comply with the first and second laws of thermodynamics. 
While we optimize the squared error $\ell(z,x) = ||z-x||_2^2$ in this work, this can easily be replaced by other loss functions.
%and we choose the loss function $\mathcal{L}(z,x) = ||z-x||_2^2$ in this work.

\begin{figure}
    \centering
    \includegraphics[width = \linewidth]{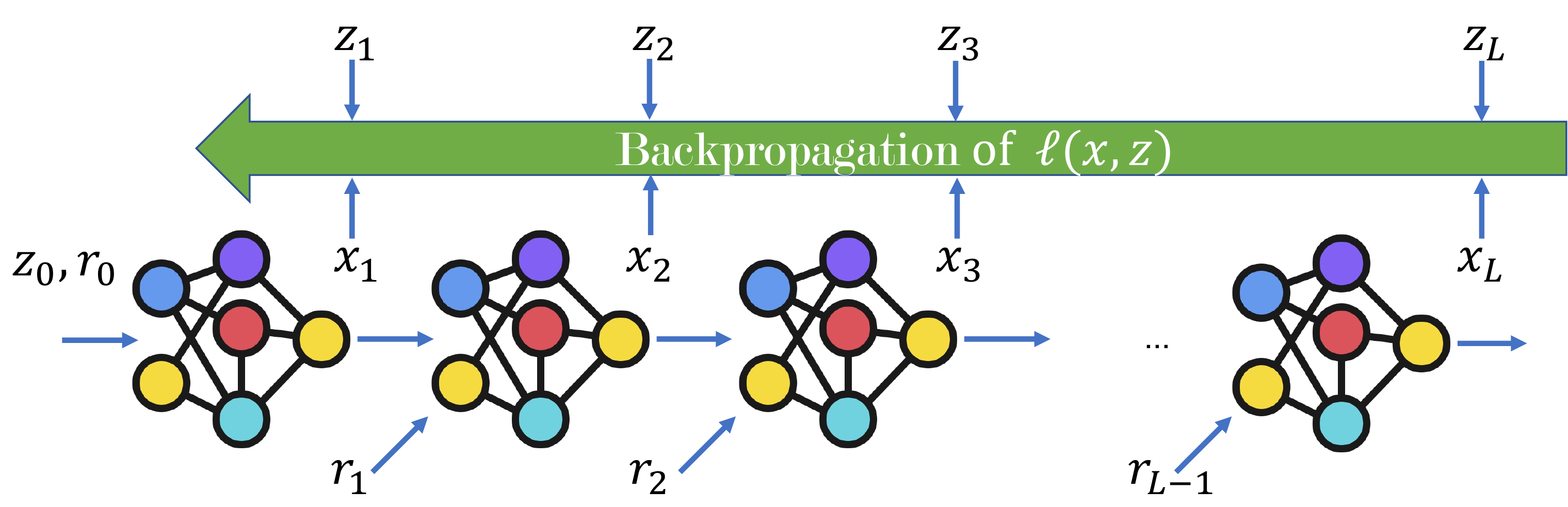}
    \caption{A pictorial description of the training procedure of the discretized PC-NODE~\eqref{eq:FE}.}
    \label{fig.backpropagation}
\end{figure}

We implement the proposed PC-NODEs using \texttt{PyTorch}, which allows us to easily propagate the inputs through the NODE and then rely on automatic BackPropagation Through Time (BPTT) \citep{werbos1990backpropagation} to run Gradient Descent (GD) on the trainable parameters,
%Depending upon the application, an additional constraint $\textcolor{blue}{g}> 0$ can be incorporated to render \eqref{PC-NODE} monotonic with respect to the inputs $u$. The training procedure can be summarized in three steps: i) forward propagate through the NODE, and calculate the cumulative loss \eqref{obj} ii) back-propagate through time and update the trainable parameters. iii) Repeat until we achieve the desired training and prediction loss. A pictorial representation of the training procedure is illustrated in the Fig.  \ref{fig.backpropagation}. We can employ the standard python packages, such as \emph{pytorch}, \emph{torchdef} in order to train the NODE \eqref{PC-NODE}. \citep{werbos1990backpropagation}
%This training procedure is
as sketched in Fig.~\ref{fig.backpropagation}. However, in general, it does not allow one to introduce constraints on the parameters directly. In particular, it cannot not guarantee that either $\textcolor{blue}{J}$ satisfies property~$(P_2)$ or $\textcolor{blue}{R}$~$(P_3)$. %has a desired sparsity pattern. 
Consequently, the next Section discusses how to ensure that these constraints are satisfied -- despite running unconstrained GD -- in the light of two illustrative examples. While this may seem counter-intuitive at first, running unconstrained GD while enforcing constraints by construction allows one to leverage the full strength of automatic GD in \texttt{PyTorch}, which naturally scales to large systems and long prediction horizons.

\begin{remark}
Besides modifying the loss function $\ell$, one can also introduce weighted penalty terms in equation~\eqref{obj}, e.g. to promote sparse solutions with $||\textcolor{blue}{J}||_1$. 
%One can easily introduce penalty terms in Equation~\eqref{obj}, for promoting sparse solutions through the addition of weighted %$L_1$
%norms, e.g. $||\textcolor{blue}{J}||_1$.
%The $L_1$ regualarization term in \eqref{obj} promotes sparsity in the interconnection matrix $\textcolor{blue}{J}$. Note that the construction of \eqref{PC-NODE} allows us to incorporate any prior or sparsity in trainable parameters while respecting the laws of thermodynamics. 
%Depending upon the application, an additional constraint $\textcolor{blue}{g}> 0$ can be incorporated to render \eqref{PC-NODE} monotonic with respect to the inputs $u$.
% We will demonstrate how to incorporate priors during training in Experiments section.  \loris{Except we don't :)}
\end{remark}

\section{Model formulations}

To demonstrate the variety of systems that can be represented with IPH dynamics, this section describes how to model thermal building dynamics and gas-piston systems and ensure properties $(P_1)$--$(P_4)$ are respected.  % with the proposed PC-NODEs.

\subsection{Thermal building dynamics}

\begin{figure}
    \centering
    \includegraphics[width = 0.6\linewidth]{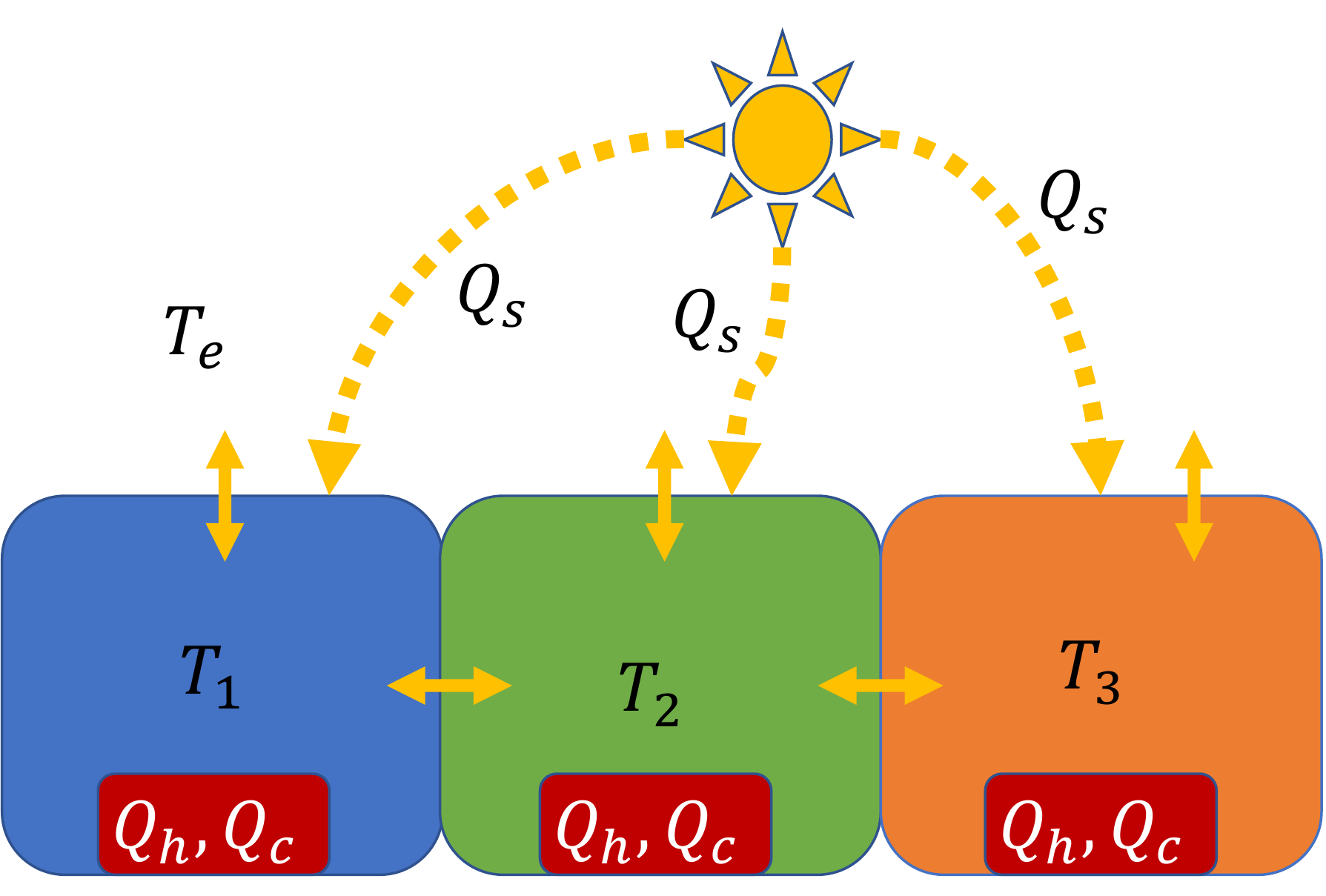}
    \caption{A pictorial description of the thermal behavior of a three-zone building, where yellow arrows represent energy flows.}
    %considered for modeling. The yellow arrows represent the flow of energy within the environment. Note that the energy from the sun is one-directional.}
    \label{fig.Building}
\end{figure}

The thermal dynamics of a building can be seen as $N$ connected thermal zones exchanging energy % and impacted by %heat gains
among themselves and with the outside, as depicted in Fig.~\ref{fig.Building}. In this work, we assume that they are additionally impacted by various heat gains from heating or cooling operations and solar irradiation. %We can hence leverage 
Inspired by the IPH formulation of heat exchangers~\citep{ramirez2013irreversible}, we model the entropy $S\in\mathbb{R}^N$ in each zone as follows
%Leveraging our prior knowledge on the thermal behavior of buildings, we can simplify the generic IPHS \eqref{PC-NODE} to model the evolution of entropy $S\in\mathbb{R}^N$ of $N$ thermal zones as \loris{REF}:
\begin{align}
\label{eq.NeuralODE}
     \dot{S} &= \textcolor{blue}{\tilde{J}}(T) \frac{\partial H(S)}{\partial S} + \textcolor{blue}{B_{e}}(T) T_e 
    + \begin{bmatrix}
    \textcolor{blue}{B_{s}} & \textcolor{blue}{B_h} & \textcolor{blue}{B_c}
    \end{bmatrix} 
    \begin{bmatrix} Q_s \\ Q_h \\ Q_c \end{bmatrix}\; ,
\end{align}
where $T\in\mathbb{R}^N$ represents the temperature in each zone. %For clarity, w
We separated the different inputs $u$, with $T_e\in\mathbb{R}$ corresponding to the ambient temperature, and $Q_s$, $Q_h$, $Q_c\in\mathbb{R}^N$ to solar, heating, and cooling gains for each zone, respectively. $\textcolor{blue}{B_s}$, $\textcolor{blue}{B_h}$, and $\textcolor{blue}{B_c}$ are $N\times N$ diagonal matrices gathering trainable %scaling
parameters reflecting the impact of these gains on the entropy of each zone. $\textcolor{blue}{B_e}(T)\in\mathbb{R}^N$ models the heat losses to the outside, with entries
\begin{equation*}
    \textcolor{blue}{B_e}(T)_i=\textcolor{blue}{\lambda_{ie}}\frac{(T_e - T_i)}{(T_i T_e)}
\end{equation*} for each zone $i$, where $\{\textcolor{blue}{\lambda_{ie}}\}_{i=1}^N$ are the trainable parameters. Finally, the skew-symmetric matrix $\textcolor{blue}{\tilde{J}}(T) \in \mathbb{R}^{N \times N}$, lumping together $R$ and $J$ in this case, is parametrized as
\begin{equation*}
     \textcolor{blue}{\tilde{J}_{ij}}(T) = -\textcolor{blue}{\tilde{J}_{ji}}(T) = \begin{cases}
    \textcolor{blue}{\lambda_{ij}}\frac{(T_j - T_i)}{(T_i T_j)} &\text{if $i$ is adjacent to $j$} \\
    0 &\text{otherwise,}
    \end{cases}    
\end{equation*}
where two zones are \textit{adjacent} if they share at least a common wall. 

Interestingly, by the definition of entropy, and recalling that the Hamiltonian $H$ represents the energy of the system, we have $\frac{\partial H(S)}{\partial S}=T$. Hence, there is no need to parametrize the partial derivatives of the Hamiltonian function in this case since they can be computed explicitly from the state of the system, as shown in Appendix~\ref{app.compTemps} (assuming a constant volume for each zone).

%since one can compute its derivatives from the state of the system, %zone temperatures from their entropy, 
%as derived in Appendix~\ref{app.compTemps} (assuming constant volume for each zone).
%directly from entropy. We defer the reader to Appendix \ref{app.compTemps} for details (assuming constant volume).

\begin{proposition}[Consistency, and monotonicity]\label{prop:cons}
The PC-NODE~\eqref{eq.NeuralODE} is consistent with the first and second laws of thermodynamics and monotonic with respect to all inputs, i.e. $T_e$, $Q_s$, $Q_h$, and $Q_c$ if the learned parameters satisfy %the ambient temperature $T_e$ and the various gains $Q_s$, $Q_h$, and $Q_c$ if
\begin{align*}
    \textcolor{blue}{B_s}, \textcolor{blue}{B_h}, \textcolor{blue}{B_c}\succeq0, \ \ \text{and} \ \  \textcolor{blue}{\lambda_{ij}}, \textcolor{blue}{\lambda_{ie}}\in\mathbb{R}_+, \ \  \forall i,j = 1,\hdots,N.
\end{align*}
\end{proposition}
\begin{pf}
See Appendix \ref{app.consitency} for a sketch of the proof and \cite{van2021energy} for more details. \hfill $\square$
\end{pf}

\begin{remark}
The dependence of $\textcolor{blue}{\tilde{J}}$ on $T$ in \eqref{eq.NeuralODE} violates property~$(P_2)$. % of system~\eqref{PC-NODE} stating that $\textcolor{blue}{\tilde{J}}$ should be constant. 
While state-dependent connection matrices break the consistency of the system with the first and second laws of thermodynamics in general~\citep{ramirez2013irreversible}, we %show how to 
show that PC-NODE~\eqref{eq.NeuralODE} remains consistent in the proof of Proposition~\ref{prop:cons}.
%these laws are still respected in the particular case of thermal building dynamics in Appendix~\ref{app.consitency}.
%on the gradient $\frac{\partial H(S)}{\partial S}= T$ destroys the symplectic structure given by the Poisson tensor associated with the structure matrix $\textcolor{blue}{J}(x)$ \citep{ramirez2013irreversible}. Nevertheless, we show how to decompose $\textcolor{blue}{J}(T)$ such that the symplectic structure remains intact.  See Appendix \ref{app.consitency} for more details. 
\end{remark}

\begin{remark}
Exploiting the linearity of the PC-NODE~\eqref{eq.NeuralODE}, one can show that it is almost equivalent to well-known Resistance-Capacitance (RC) architectures. The latter model the energy of each zone instead of their entropy, but both quantities are linked by definition since 
%which model the energy of the system instead of its entropy. % \citep{van2014port}. %The main difference indeed comes from the fact that \eqref{eq.NeuralODE} models the entropy of the system to exploit the underlying Hamiltonian structure, while RC methods model the energy in each zone. 
%However, both are linked by definition since 
$dS = \frac{dH}{T}$. Multiplying \eqref{eq.NeuralODE} by the temperature of each zone, one can hence recover an energy model of the building, but with the training %gain 
parameters in $\textcolor{blue}{B_s}(T)$, $\textcolor{blue}{B_h}(T)$, and $\textcolor{blue}{B_c}(T)$ depending on the corresponding zone temperatures. Since the latter can be considered as roughly constant (in Kelvin), this is indeed similar to classical RC models.
%However, both are linked by definition dince $dS = \frac{dH}{T}$. Multiplying \eqref{eq.NeuralODE} by $T$ in each dimension, one can hence recover a typical RC model, but with the parameters in $\textcolor{blue}{B_s}$, $\textcolor{blue}{B_h}$, and $\textcolor{blue}{B_c}$ depending on the corresponding zone temperature. Since zone temperatures can be considered as roughly constant (in Kelvin), this will approach classical RC modeling techniques.
\end{remark}

\subsection{Gas Piston system}

\begin{figure}
    \centering
    \includegraphics[height = 25mm]{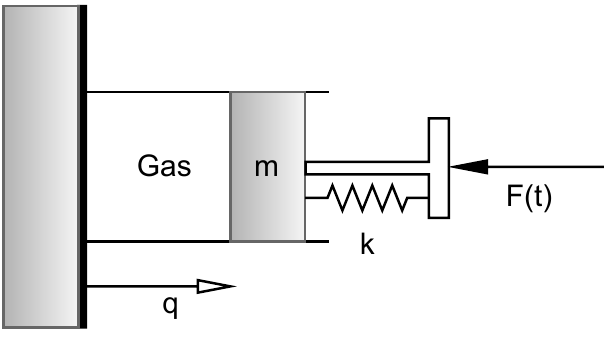}
    \caption{Sketch of the gas piston system.}
    \label{fig:gasPiston}
\end{figure}

Consider a typical gas piston system, as depicted in Fig.~\ref{fig:gasPiston}, where the piston is subject to friction, influenced by an external force $F(t)=u$, and dampened by a spring. We define the state of the system as $x=[S,V,q,p]^\top$, where $S$ is entropy and $V$ the volume of the gas, and $q$, $p$ are the position and momentum of the piston, respectively. %A sketch of the system is given in Fig. \ref{fig:gasPiston}. 
Inspired by \citet{ramirez2013modelling}, the system can be described by the following nonlinear IPH dynamics
\begin{align}
\dot{x} &= \left[\textcolor{blue}{R}\left(x, \frac{\partial S}{\partial x}, \frac{\partial H(x)}{\partial x}\right) J_0 + \textcolor{blue}{J_1} \right] \frac{\partial \textcolor{blue}{H}(x)}{\partial x} + G u \;, \label{equ:gas piston}
\end{align}
\begin{align*}
J_0 &= \begin{bmatrix}
0 & 0 & 0 & 1 \\
0 & 0 & 0 & 0 \\
0 & 0 & 0 & 0 \\
-1 & 0 & 0 & 0 
\end{bmatrix},
&\textcolor{blue}{J_1} &= \begin{bmatrix}
0 & 0 & 0 & 0 \\
0 & 0 & 0 & \textcolor{blue}{\alpha} \\
0 & 0 & 0 & \textcolor{blue}{\beta} \\
0 & \textcolor{blue}{-\alpha} & \textcolor{blue}{-\beta} & 0 
\end{bmatrix}, \\
\frac{\partial \textcolor{blue}{H}(x)}{\partial x} &=  \begin{bmatrix}
\textcolor{blue}{T, -P, Kq, v} 
\end{bmatrix}^\top,
&G &= \begin{bmatrix}
0, 0, 0, 1 
\end{bmatrix}^\top \;,
\end{align*}
where $\textcolor{blue}{R}(x, \frac{\partial S}{\partial x}, \frac{\partial \textcolor{blue}{H}(x)}{\partial x}) = \textcolor{blue}{\frac{\mu v}{T}}$, $T$ is the temperature and $P$ the pressure of the gas, $K$ the spring constant, and $v=\frac{p}{m}$ represents the speed of the piston with mass $m$ %, area $A$, 
and friction coefficient $\mu$. 

Since the entropy is a state of the system, $\frac{\partial S}{\partial x} = [1, 0, 0, 0]^\top$, which implies that property~$(P_2)$ becomes
\begin{align}
    \textcolor{blue}{R}\left(x, \frac{\partial S}{\partial x}, \frac{\partial \textcolor{blue}{H}(x)}{\partial x}\right) 
    &= \textcolor{blue}{\gamma}\left(x, \frac{\partial \textcolor{blue}{H}(x)}{\partial x}\right) \frac{\partial S}{\partial x}^\top J_0 \frac{\partial \textcolor{blue}{H}(x)}{\partial x} \nonumber \\
    %&= \gamma\left(x, \frac{\partial H(x)}{\partial x}\right) \frac{\partial H(x)}{\partial x_4} \\
    &= \textcolor{blue}{\gamma}\left(x, \frac{\partial \textcolor{blue}{H}(x)}{\partial x}\right) \frac{\partial \textcolor{blue}{H}(x)}{\partial p}\; .
\end{align}
The function $\textcolor{blue}{R}$ is thus well-defined and can be derived from $\textcolor{blue}{\gamma}$ and $\textcolor{blue}{H}$. To showcase the flexibility of the proposed PC-NODEs, we assume the Hamiltonian to be unknown and parametrize it as a single-layer NN with the form
\begin{align}
    \textcolor{blue}{H}(x;\textcolor{blue}{\theta}) &= \log \left[ \cosh (\textcolor{blue}{K}x + \textcolor{blue}{b})\right]^\top \mathds{1}_n \;, 
\end{align}
where $\mathds{1}_n$ represents a column vector with $n$ elements equal to $1$, and $\textcolor{blue}{\theta}=\{\textcolor{blue}{K},\textcolor{blue}{b}\}$.  Such an architecture is chosen for its elegance because it allows us to compute the required partial derivatives in closed form\citep{galimberti2021hamiltonian}
\begin{align}
    \frac{\partial \textcolor{blue}{H}(x;\textcolor{blue}{\theta})}{\partial x} &= \textcolor{blue}{K}^\top \tanh (\textcolor{blue}{K}x + \textcolor{blue}{b}) \;.
\end{align}
We parametrize $\textcolor{blue}{\gamma}$ as a single layer NN $\textcolor{blue}{\gamma}:\mathbb{R}^8\to\mathbb{R}^+$, where positivity is obtained by feeding the output through a sigmoid function, which is sufficient to ensure property~$(P_2)$. Finally, we assume the sparsity pattern of $\textcolor{blue}{J_1}$ to be known, but not its parameters $\{\textcolor{blue}{\alpha},\textcolor{blue}{\beta}\}$%\footnote{With a slight abuse of notation, we assume the constant number $1$ to be unknown and learn it, similarly to $A$.}
, to demonstrate how prior knowledge might be incorporated into the learning process\footnote{In the true system, $\alpha$ is the area of the piston and $\beta=1$.}. %\loris{Need to explicitly mention "these and these are the learned parameters"? (same for building)}
%All the parameters $\{\textcolor{blue}{\theta}, \textcolor{blue}{\beta}, \textcolor{blue}{\eta}\}$ are identified from sampled trajectories
%After discretizing the system with the FE method, all the parameters $\{\textcolor{blue}{\theta}, \textcolor{blue}{\beta}, \textcolor{blue}{\eta}\}$ can be learned from sample trajectories by minimizing the cost function in \eqref{obj}.

\begin{remark}
Although PC-NODE~\eqref{equ:gas piston} is slightly different from the generic representation in~\eqref{PC-NODE}, the key system properties are still conserved. Indeed, one can always decompose the product between $R$ and $J$ in a sum of products without violating the first and second laws of thermodynamics as long as each term respects condition~\eqref{equ:R} and the skew-symmetry of $J$. See the proof of Proposition~\ref{prop:cons} for more details.
\end{remark}

\begin{remark}\label{rem:entropy}
Since we assume no thermal exchanges between the gas and the ambient air, according to the second law of thermodynamics, the entropy of the gas can never decrease. 
\end{remark}

\section{Applications and results}
\label{sec:results}

This section presents the results obtained by fitting the two PC-NODEs described in the previous Section on real-world measurements for building thermal dynamics, and on simulated data for the gas piston system\footnote{The code and data can be found on \url{https://gitlab.nccr-automation.ch/loris.dinatale/pc-node}.}.

\subsection{Building thermal dynamics}

For the first application, we aim to identify the temperature dynamics of a residential apartment in NEST, a vertically integrated district in Duebendorf, Switzerland \citep{nest}. It is composed of two bedrooms separated by a living room, as sketched in Fig.~\ref{fig.Building}, where we neglected the impact of the two small bathrooms and processed the data similarly to~\citet{di2022physically}, including the computation of solar gains from horizontal irradiation measurements. Overall, three years of measurements of the temperature and solar gains in each zone, the respective heating and cooling powers, and the ambient temperature are available. %The latter were computed from solar irradiation on a flat surface measurements \citep[App. A]{di2022physically}, and turn out to be identical for all the three zones since they share the same orientation. 
The dataset has a sampling time of \SI{15}{\minute} and was split into a training and a validation set %which do not overlap, and each 
containing %possibly overlapping 
time series of measurements %without missing values. %. Note that the time series were
%Each time series is 
truncated after three days, i.e. $288$ steps, to alleviate the computational burden. %and avoid too much overlapping.

\begin{figure}
    \centering
    \includegraphics[width=\linewidth]{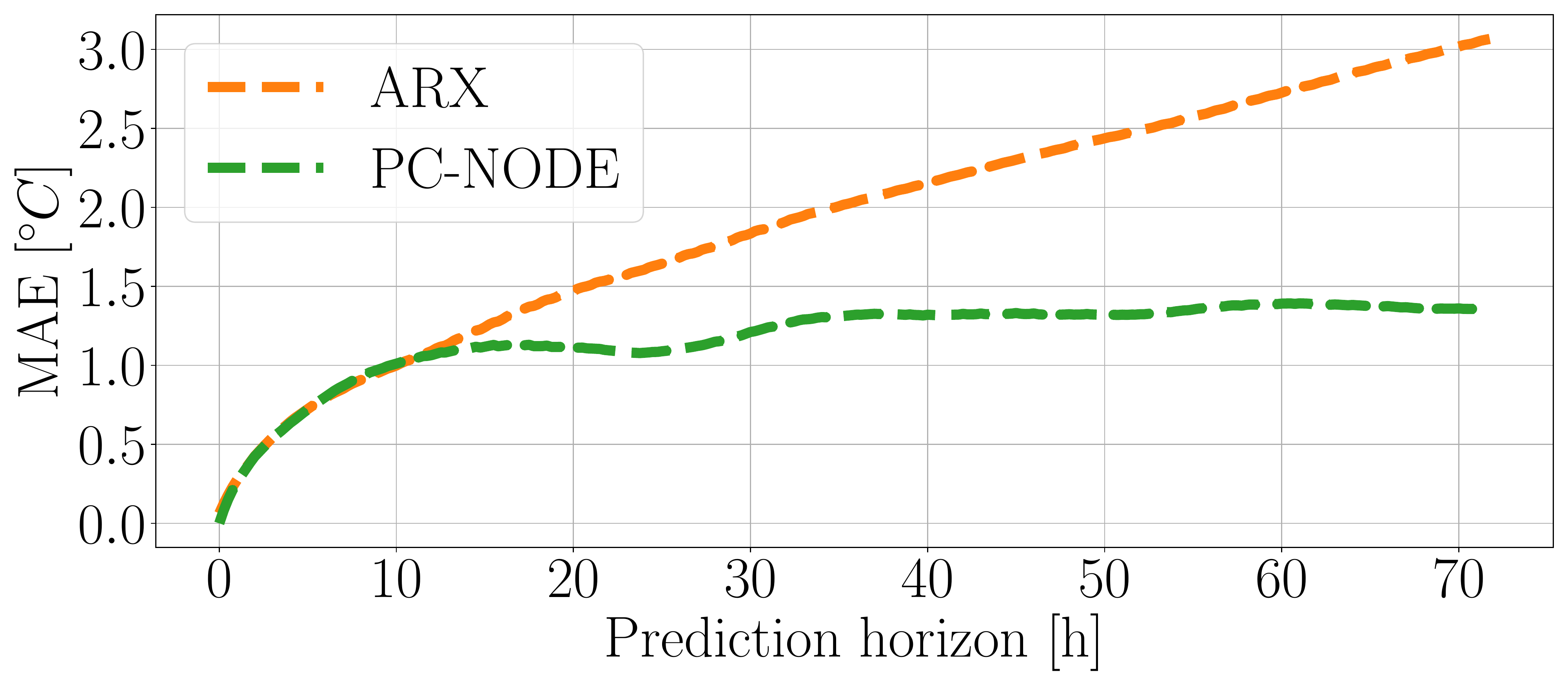}
    \caption{MAE of the ARX model and PC-NODE over the prediction horizon averaged over the three zones and the validation time series. }
    \label{fig.BuildingErrors}
\end{figure}

To investigate the performance of the learned model, we analyze its accuracy over more than $750$ sequences of three days of validation data. Averaged over the three zones and all the time series, the Mean Absolute Error (MAE) propagation over the \SI{72}{\hour} horizon is depicted in Fig.~\ref{fig.BuildingErrors}, where the absolute error is computed as $||T_k - T(k)||_1$ for $T_k$ the model prediction and $T(k)$ the measured temperature at each time step $k$. Since PC-NODE~\ref{eq.NeuralODE} is linear, we also plot the performance of a classical linear ARX model with $12$ lag terms for reference, where the number of lags was tuned empirically and the parameters fitted to the data through least squares identification, similarly to~\cite{merema2022demonstration}, for example.
%As reference, we also plot the performance of a classical ARX model with $12$ lag terms, where the number of lags was tuned empirically and the parameters fitted to the data through least squares identification. %, i.e. looking at the last \SI{3}{\hour} of data to predict the next temperature in each zone. 

As can be readily seen, thanks to the underlying physics captured by the Hamiltonian framework, the PC-NODE is able to fit the data significantly better, especially over long horizons. Indeed, it seems to be less prone to compounding errors: it improves the accuracy by $38.9\%$ compared to the ARX on average over the entire prediction horizon, but this proportion rises to $55.8\%$ at the end of the \SI{72}{\hour}-long horizon.

%\begin{figure}
%    \centering
%    \includegraphics[width=\linewidth]{building_trajectories.pdf}
%    \caption{Effect of heating and cooling, March $10$--$13$. The same power sequence is applied in each zone (bottom plot), either heating (red), cooling (blue) or turning everything off (black), and the three top plots depict the resulting temperatures in each zone.}
%    \label{fig.BuildingConsitency}
%\end{figure}

\begin{figure}
    \centering
    \includegraphics[width=\linewidth]{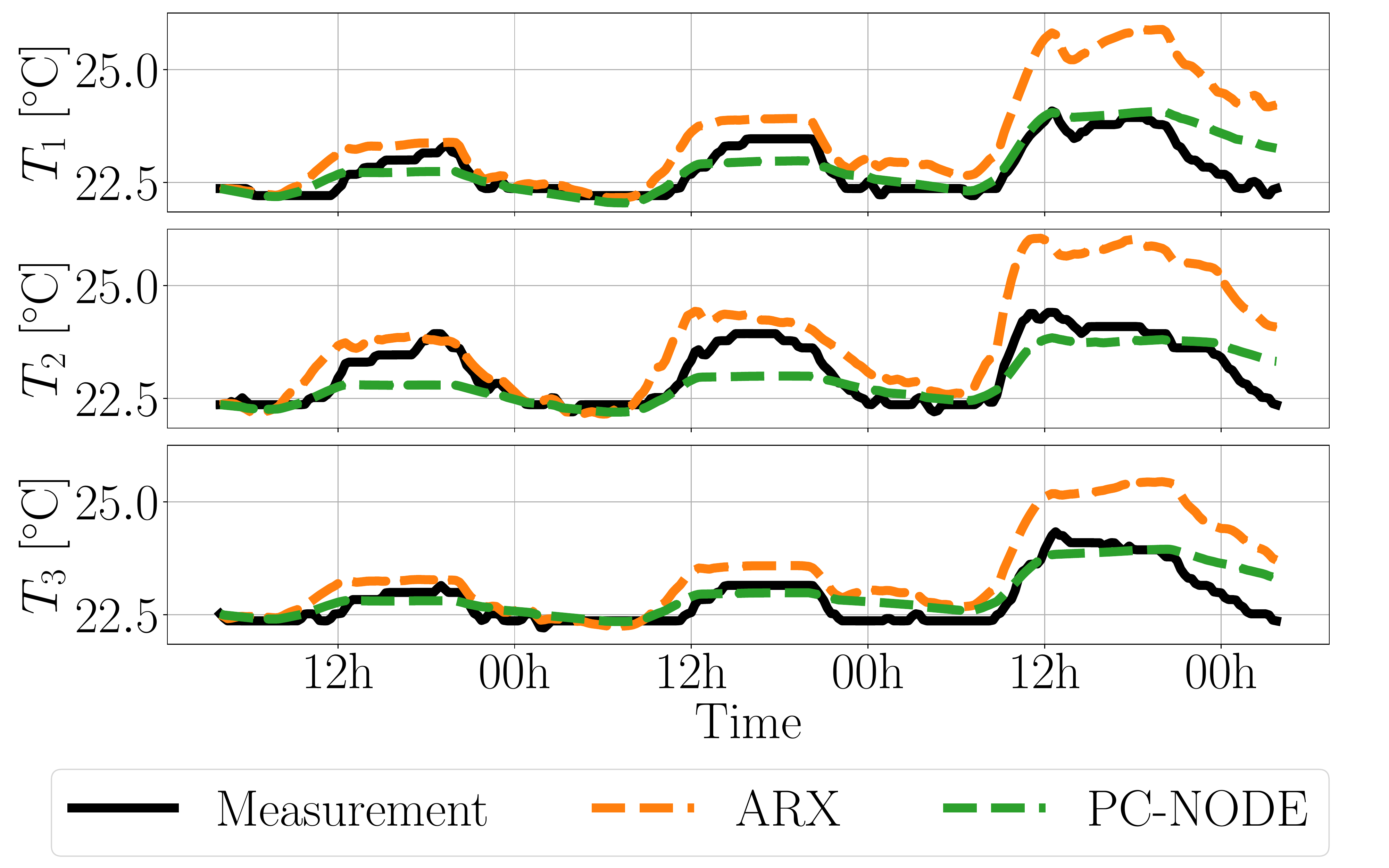}
    \caption{August $18$--$20$, $2021$: Temperature predictions of the PC-NODE and ARX model on a sampled validation trajectory, compared to the true measurements.}
    \label{fig.BuildingTrajectory}
\end{figure}

% To illustrate the thermodynamic consistency of the proposed PC-NODE, Fig.~\ref{fig.BuildingConsitency} depicts the reaction of the model to different heating and cooling inputs. As expected, heating the building consistently leads to higher temperatures down the horizon compared to when it is turned off, while cooling results in lower temperatures. This showcases the power of the proposed IPH framework, which ensures the underlying physics is respected by construction, independently of the parameter values identified from data. %\loris{Slightly tircky: ARX looks fine as well here.}

In order to provide a visual comparison of the behavior of both models, we plot their temperature predictions over a sampled \SI{72}{\hour}-long trajectory in August $2021$ in Fig.~\ref{fig.BuildingTrajectory}. %, with the ground truth measurements as a reference. 
This figure hints that the ARX model is more sensitive to the various external gains, having a tendency to overestimate their impact. This can for example be observed towards the end of the horizon in Fig.~\ref{fig.BuildingTrajectory}, just before noon: when the sun rises, increasing the temperature of the building, the ARX cannot accurately capture this behavior, contrary to the PC-NODE. While only one sampled trajectory is presented in this paper for the sake of brevity, these effects generally hold across the validation dataset and explain the better performance of the PC-NODE. 

%To compare the temperature predictions on the validation data, we plot the evolution of the temperatures in all three rooms for a horizon of 72 hours in Fig \ref{fig.BuildingTrajectory}. We can see that ARX model performs poorly for longer prediction horizon, whereas, thanks to physical-consistency, PC-NODE is performing reasonably well. 

\subsection{Gas piston system}

For the second application, we generated a synthetic dataset of $10'000$ samples from system~\eqref{equ:gas piston} using the \texttt{odeint} framework from \texttt{scipy}. The parameters are provided in Appendix~\ref{app:params} and the gas temperature has been computed as presented in Appendix~\ref{app.compTemps}. %Note that this requires access to the gas temperature and pressure: the former is computed as detailed in Appendix~\ref{app.compTemps}, and the latter can then be derived from the ideal gas law $PV=nRT$. 
To generate the training data, we then added Gaussian noise on each dimension $d$ of the state, with $\epsilon_d\sim\mathcal{N}(0,0.2\sigma_{x_d})$, where $\sigma_{x_d}$ corresponds to the standard deviation of the $d$th dimension of $x$. %, while noiseless data is used for validation.
Similarly to the previous application, the data was split into chunks of $250$ steps to alleviate the computational burden. It was additionally normalized to ease the training of the NNs used in PC-NODE~\eqref{equ:gas piston}.

\begin{figure}
    \centering
    \includegraphics[width=\linewidth]{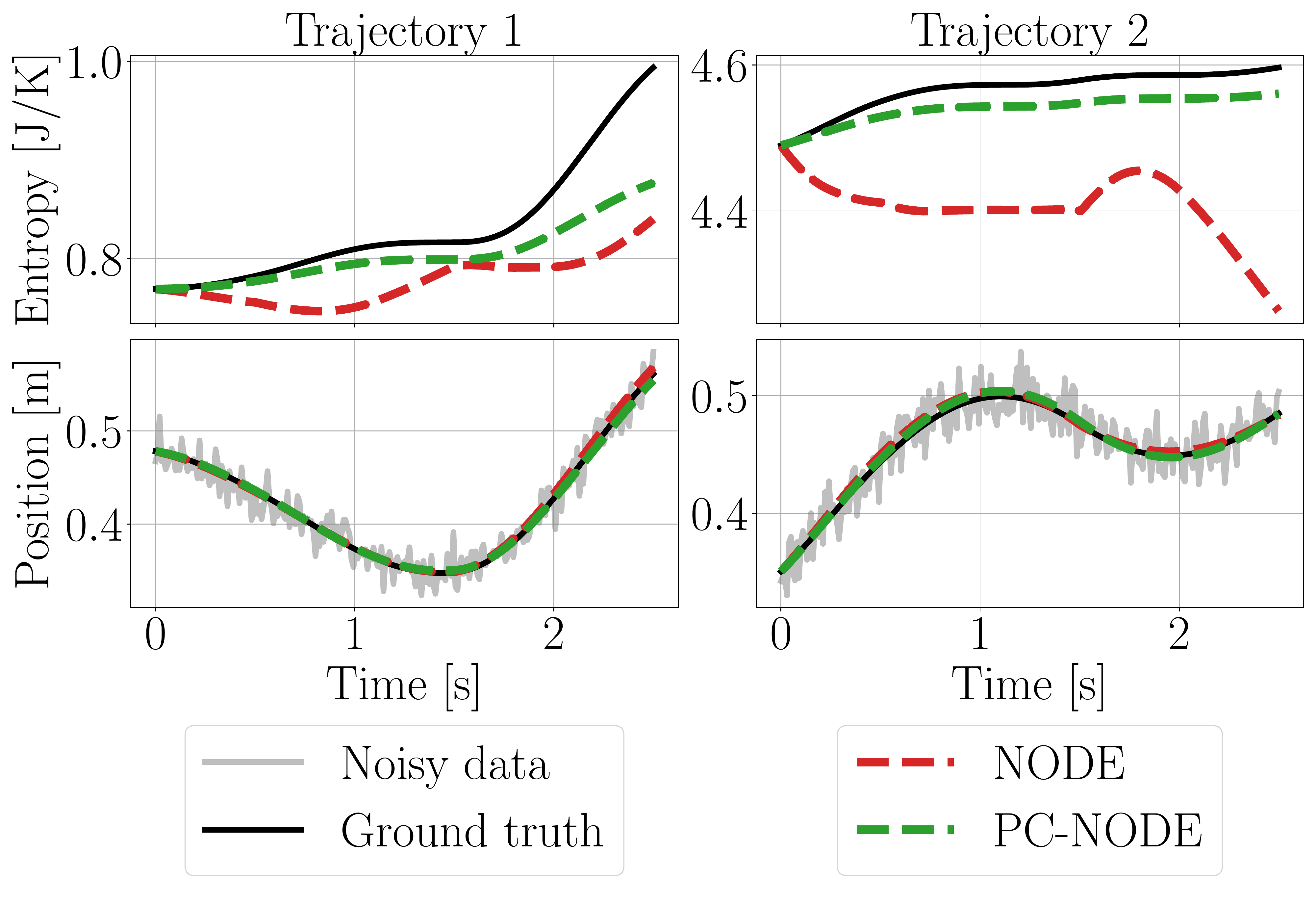}
    \caption{Sampled trajectories of the piston position and the entropy of the gas ($\times 10^3$) over time.} % The noisy data can be found in shaded grey for the position, the ground truth in black, and the predictions of the PC-NODE and classical NN in dashed green and red, respectively.}
    \label{fig.GasPiston}
\end{figure}

Despite not having access to the true Hamiltonian function and learning it as an NN from data, and even in the presence of white noise, PC-NODE~\eqref{equ:gas piston} can accurately recover the position of the system, as pictured in Fig.~\ref{fig.GasPiston} (bottom) for two sampled trajectories. However, a vanilla NODE, i.e. $\dot{x} = f_\theta(x)$~\citep{chen2018neural}, where $f_\theta$ is a NN with two hidden layers of $32$ neurons each, is also able to fit this data very well. On the other hand, the evolution of the entropy is more challenging to capture, as pictured on the top of Fig.~\ref{fig.GasPiston}, where we %scaled it and 
removed the noisy data for clarity. %because the magnitude of the changes %from one step to the other is much smaller. 
In that case, the PC-NODE clearly outperforms the vanilla NODE. %, more accurately representing the evolution of entropy. 
Remarkably, the NODE sometimes predicts a decrease in entropy, which is inconsistent with the underlying physics (Remark~\ref{rem:entropy}) %with the second law of thermodynamics 
and does not happen with the PC-NODE.

%While not depicted here for the sake of brevity, similar results were observed on the other three variables, i.e. the entropy, volume, and momentum of the system.

\section{Concluding Remarks}
This work proposed PC-NODEs, NODEs endowed with IPH dynamics, to identify multi-physics systems from data. PC-NODEs are consistent with the first and second laws of thermodynamics by construction if simple conditions are respected, which allows one to rely on automatic BPTT to identify their parameters. Leveraging prior knowledge of these systems, the proposed framework demonstrated promising performance on both a thermal building modeling and a gas piston case study.

We believe these results can pave the way for large-scale distributed data-driven models with physical consistency guarantees. %, and future work could apply PC-NODEs to other applications.

%We proposed NODEs endowed with irreversible Hamiltonian dynamics for modeling thermodynamic systems from data. They are consistent with the first and second laws of thermodynamics by structure. Therefore, enable one to perform almost free parametrization for trainable parameters of NNs. The promising performance of these NODEs is demonstrated by the real-world modeling of buildings and a gas piston system. 
%Future efforts will be devoted to incorporating distributed modelling and control.

% \begin{ack}
% Place acknowledgments here.
% \end{ack}
% \bibliographystyle{unsrtnat}
\bibliography{ifacconf.bib}             % bib file to produce the bibliography
                                                     % with bibtex (preferred)
                                                   
% \begin{thebibliography}{23pt}  % you can also add the bibliography by hand

%\bibitem[Able(1956)]{Abl:56}
%B.C. Able.
%\newblock Nucleic acid content of microscope.
%\newblock \emph{Nature}, 135:\penalty0 7--9, 1956.

%\bibitem[Able et~al.(1954)Able, Tagg, and Rush]{AbTaRu:54}
%B.C. Able, R.A. Tagg, and M.~Rush.
%\newblock Enzyme-catalyzed cellular transanimations.
%\newblock In A.F. Round, editor, \emph{Advances in Enzymology}, volume~2, pages
%  125--247. Academic Press, New York, 3rd edition, 1954.

%\bibitem[Keohane(1958)]{Keo:58}
%\newblock \emph{Power and Interdependence: World Politics in Transitions}.
%\newblock Little, Brown \& Co., Boston, 1958.

%\bibitem[Powers(1985)]{Pow:85}
%\newblock Is there a way out?
%\newblock \emph{Harpers}, pages 35--47, June 1985.

%\bibitem[Soukhanov(1992)]{Heritage:92}
%\newblock \emph{{The American Heritage. Dictionary of the American Language}}.
%\newblock Houghton Mifflin Company, 1992.

%\end{thebibliography}

\appendix
\section{Temperature computation}   
\label{app.compTemps}% Each appendix must have a short title.
By definition, the energy of a mass $m$ of air can be described as $E(T) = mc(T)T$, where $c$ is the specific heat capacity of air and $T$ is temperature. Assuming we deal with an ideal gas, we also know that $PV=nRT$. This allows us to rewrite the time derivative of entropy, by definition satisfying
%In this appendix, we provide a method for computing temperatures $T$ from the entropy $S$. We exploit the fact, that the energy of a mass of air depends linearly on the heat capacity and the current temperature, i.e. $E(T) = mc(T)T$ and we assume ideal gas equation, that is, $PV = nRT$. By definition, the time derivative of entropy is defined as
\begin{align*}
    \frac{dS}{dt} &= \frac{1}{T}\frac{dE}{dt} + \frac{P}{T}\frac{dV}{dt}  \\
                  &= \frac{1}{T}\frac{d}{dt}(mc(T)T) + \frac{nRT}{VT}\frac{d}{dt}V \;.
\end{align*}
Since the temperature does not change abruptly, we can assume a constant heat capacity $c(T) \approx c$ and obtain
\begin{align}
\label{eq.dotEntropy}
              \frac{dS}{dt}    \approx mc\frac{\frac{dT}{dt}}{T} + nR\frac{\frac{dV}{dt}}{V} 
                  = mc\frac{d\ln{T}}{dt} + nR\frac{d\ln{V}}{dt} \;.
\end{align}
Integrating \eqref{eq.dotEntropy} on both sides from an initial time $t_i$ to a final time $t_f$, we get
\begin{align*}
\int_{t_i}^{t_f}{\frac{dS}{dt}dt} &= mc \int_{t_i}^{t_f}{\frac{d\ln{T}}{dt}dt} + nR \int_{t_i}^{t_f}{\frac{d\ln{V}}{dt}dt}, 
\end{align*}
leading to
\begin{align*}
    S(t_f) - S(t_i)           &= mc \ln{\frac{T(t_f)}{T(t_i)}} + nR \ln{\frac{V(t_f)}{V(t_i)}} \\
      &= \ln{\left[\left(\frac{T(t_f)}{T(t_i)}\right)^{mc}\left(\frac{V(t_f)}{V(t_i)}\right)^{nR}\right]} \;. 
\end{align*}
We can thus compute the final temperatures as
\begin{align*}
\label{eq:tempUpdate}
    T(t_f) = \left[\exp \left({\frac{S(t_f) - S(t_i)}{mc}} \right)\left(\frac{V(t_f)}{V(t_i)}\right)^{\frac{-nR}{mc}}\right]T(t_i) \;.
\end{align*}

\section{Proof of Proposition~\ref{prop:cons}}
\label{app.consitency}
Let us define $\mathcal{E}=\{(i,j)| \text{ Zones }i\text{ and }j \text{ are adjacent}\}$, the set of connections between the thermal zones, and consider the following decomposition of %the skew-symmetric matrix 
$\textcolor{blue}{\tilde{J}}(T)$: 
\begin{align*}
    \textcolor{blue}{\tilde{J}}(T) = \sum_{k\in \mathcal{E}} \textcolor{blue}{R_{k}}(T) \textcolor{blue}{\mathcal{J}_{k}} \; ,
\end{align*}
where $\textcolor{blue}{R_{k}}:\mathbb{R}^N \mapsto \mathbb{R}$, $\textcolor{blue}{R_{k}}(T) = \textcolor{blue}{\lambda_{ij}}\frac{(T_j - T_i)}{(T_i T_j)}$ %$k$ the edge between zones $i$ and $j$,
and $\textcolor{blue}{\mathcal{J}_{k}}$ is an $N \times N$ constant skew-symmetric matrix with zeroes everywhere, except $(J_{k})_{ij}=-(J_{k})_{ji} = 1$, for $k=(i,j)$. 

Then, for $T_e, Q_s, Q_h, Q_c \equiv 0$, we have:
\begin{align*}
    \frac{ d H}{dt} &= \frac{\partial H(S)}{\partial S}^\top \dot{S} =  \frac{\partial H(S)}{\partial S}^\top\left[ \sum_{k\in \mathcal{E}} \textcolor{blue}{R_{k}}(T) \textcolor{blue}{\mathcal{J}_k}\right] \frac{\partial H(S)}{\partial S} \\
     &=  \sum_{k\in \mathcal{E}} \left[ \frac{\partial H(S)}{\partial S}^\top \textcolor{blue}{R_{k}}(T) \textcolor{blue}{\mathcal{J}_k} \frac{\partial H(S)}{\partial S}\right] = 0\;,
\end{align*}
since each term of the sum is zero, as in equation~\eqref{equ:conservation energy}, because each $\textcolor{blue}{\mathcal{J}_k}$ is now constant and each $\textcolor{blue}{R_{k}}(T)$ satisfies condition ($P_2$). This proves the required conservation of energy of the system. 

In order to verify the irreversible creation of total entropy $S_{t}$ of the system, we note that for %each $k=(i,j)$,
$T_e, Q_s, Q_h, Q_c \equiv 0$: %, for $T_e = Q_s = Q_h = Q_c \equiv 0$: 
\begin{align*}
    \textcolor{blue}{R_{k}}(T) = \left(\frac{\textcolor{blue}{\lambda_{ij}}}{T_i T_j}\right) \{S_{t},H\}_{\textcolor{blue}{\mathcal{J}_k}}\; ,
\end{align*}
as in the case of two heat exchangers~\citep{ramirez2013irreversible}. Since the total entropy is the sum of the entropy in each zone $d$, we get
\begin{align*}
    \dot{S}_{t} &= \sum_{d=1}^n(\dot{S})_d =  \sum_{d=1}^n \left(\left[ \sum_{k \in \mathcal{E}} \textcolor{blue}{R_{k}} \textcolor{blue}{\mathcal{J}_k} \right]\frac{\partial H(S)}{\partial S} \right)_d \\
    &= \sum_{k \in \mathcal{E}} \textcolor{blue}{R_{k}} \sum_{d=1}^n \left( \textcolor{blue}{\mathcal{J}_k} \frac{\partial H}{\partial S} \right)_d 
    = \sum_{k \in \mathcal{E}} \textcolor{blue}{R_{k}} \left( \mathds{1}_n^\top \textcolor{blue}{\mathcal{J}_k} \frac{\partial H}{\partial S} \right) \\
    &= \sum_{k \in \mathcal{E}} \textcolor{blue}{R_{k}} \left(\frac{\partial S_t}{\partial S}^\top \textcolor{blue}{\mathcal{J}_k} \frac{\partial H}{\partial S} \right) 
    =  \sum_{k \in \mathcal{E}} \frac{\textcolor{blue}{\lambda_{ij}}}{T_i T_j}  \{S_t,H\}_{\textcolor{blue}{\mathcal{J}_k}}^2 \geq 0\; ,
\end{align*}
since $\frac{\partial S_t}{\partial S} = \mathds{1}_n$ by definition, and the inequality holds if all $\{\textcolor{blue}{\lambda_{ij}}\}_{(i,j)\in\mathcal{E}}$ are positive since temperatures are positive.
%where we used the fact that the entropy is the state of the system, i.e. $\frac{\partial S}{\partial S_d} = e_d$, where $e_d$ is the natural basis, and the last inequality holds if all $\{\textcolor{blue}{\lambda_{ij}}\}_{(i,j)\in\mathcal{E}}$ are positive. Since the total entropy of the system is the sum of the entropy in each zone $d$, this concludes the proof of irreversibly entropy creation.

Finally, if all the input matrices $\textcolor{blue}{B_e}, \textcolor{blue}{B_s}, \textcolor{blue}{B_h}$, and $\textcolor{blue}{B_c}$ are positive definite, monotonicity follows from the fact that PC-NODE \eqref{eq.NeuralODE} is affine in input by construction. \hfill $\square$

\section{Gas piston parameters}
\label{app:params}

The gas piston system was simulated from $T(0)=\SI{290}{\kelvin}$, $x(0)=[0,0.001,0.3,0]^\top$, with $m=\SI{5}{\kilo\gram}$, area $\alpha=\SI{0.033}{\meter\squared}$, $\beta = 1$, $\mu=1$, and $K=\SI{10}{\newton\per\meter}$, and the sampling time $h=\SI{0.01}{\second}$.

\end{document}